\documentclass[11pt,a4paper]{article}
\usepackage[hyperref]{emnlp2020}
\usepackage{times}
\usepackage{latexsym}

\usepackage{amsmath,amssymb,mathtools}
\usepackage{tikz}
\usetikzlibrary{positioning}
\usepackage[framemethod=TikZ]{mdframed}
\mdfsetup{%
 middlelinewidth=0,
 backgroundcolor=gray!15,
 roundcorner=8pt
}

\usepackage{lingmacros}
\usepackage{bm}
\usepackage{caption}
\usepackage{subcaption}
\usepackage{booktabs}
\usepackage{amsfonts}
\usepackage{CJKutf8}
\usepackage{cleveref}
\usepackage{amsmath, amsfonts}
\usepackage{xcolor}
\usepackage{soul}
\usepackage{comment}
\usepackage{threeparttable}
\usepackage{color}
\usepackage{url}
\usepackage{graphicx}
\usepackage{here}
\usepackage[linesnumbered,ruled,vlined]{algorithm2e}
\usepackage{multirow}
\usetikzlibrary{positioning,calc,shapes.geometric,backgrounds,arrows.meta}
\usepackage{amsthm}
\usepackage{array}
\usepackage{colortbl}
\usepackage[whole]{bxcjkjatype}
\usepackage{pifont}
\usepackage{setspace}
\usepackage{float}

\newcommand{\figcaption}[1]{\def\@captype{figure}\caption{#1}}
\newcommand{\tblcaption}[1]{\def\@captype{table}\caption{#1}}
\newcommand{\fscore}{\ensuremath{\mathrm{F}_{0.5}}}

\newcommand{\truseteddata}{\ensuremath{\mathcal{\hat{D}}}}
\newcommand{\noisydata}{\ensuremath{\mathcal{{D}}}} %

\newcounter{magicrownumbers}
\newcommand\rownumber{\scriptsize{\stepcounter{magicrownumbers}\arabic{magicrownumbers} :}}

\newcommand{\rx}{\scriptsize{$\bm{X}$ : } } %
\newcommand{\ry}{\scriptsize{$\bm{Y}$ : } } %
\newcommand{\ryy}{\scriptsize{$\bm{Y}^{\prime}$: } } %

\newcolumntype{H}{@{}>{\lrbox0}l<{\endlrbox}}
\newcolumntype{+}{>{\global\let\currentrowstyle\relax}}
\newcolumntype{^}{>{\currentrowstyle}}

\usepackage{microtype}

\aclfinalcopy %
\def\aclpaperid{347} %

\title{A Self-Refinement Strategy for Noise Reduction \\ in Grammatical Error Correction}

\author{Masato Mita$^{1,2}$
Shun Kiyono$^{1,2}$
Masahiro Kaneko$^{3,1}$
Jun Suzuki$^{2,1}$
Kentaro Inui$^{2,1}$\\
 $^{1}$RIKEN Center for Advanced Intelligence Project \\
 $^{2}$Tohoku University \\
 $^{3}$Tokyo Metropolitan University \\
 {\tt \{masato.mita, shun.kiyono\}@riken.jp}\\
 {\tt kaneko-masahiro@ed.tmu.ac.jp} \\
 {\tt \{jun.suzuki, inui\}@ecei.tohoku.ac.jp}
 }

\date{}

\begin{document}
\maketitle
\begin{abstract}
Existing approaches for grammatical error correction (GEC) largely rely on supervised learning with manually created GEC datasets.
However, there has been little focus on verifying and ensuring the quality of the datasets, and on how lower-quality data might affect GEC performance. 
We indeed found that there is a non-negligible amount of ``noise'' where errors were inappropriately edited or left uncorrected. 
To address this, we designed a self-refinement method where the key idea is to denoise these datasets by leveraging the prediction consistency of existing models, and outperformed strong denoising baseline methods. 
We further applied task-specific techniques and achieved state-of-the-art performance on the CoNLL-2014, JFLEG, and BEA-2019 benchmarks. 
We then analyzed the effect of the proposed denoising method, and found that our approach leads to improved coverage of corrections and facilitated fluency edits which are reflected in higher recall and overall performance.
\end{abstract}

\section{Introduction}
\label{sec:Intro}

Grammatical error correction (GEC) is often considered a variant of machine translation (MT) ~\citep{brockett-etal-2006-correcting,junczys:2018:NAACL} due to their structural similarity--``translating" from source ungrammatical text to target grammatical text.
At present, several neural encoder-decoder (EncDec) approaches have been introduced for this task and have achieved remarkable results~\citep{chollampatt:2018:AAAI, zhao2019improving, kiyono-etal-2019-empirical}.
EncDec models tend to further improve in performance with increasing data size~\citep{koehn:2017:NMT,sennrich:2019:ACL}, however, this is not necessarily true in GEC.
\begin{table}[!t]
\centering
\begin{tabular}{l}
\toprule
1 : Errors are inappropriately edited   \\ \midrule
Source: {\it I want to \textbf{discuss about} the education.}      \\
Target: {\it I want to \textbf{discuss of} the education.}                    \\
\midrule
2 : Errors are left uncorrected \\  \midrule
Source: {\it We \textbf{discuss about} our sales target.}   \\
Target: {\it We \textbf{discuss about} our sales target.}      \\ \bottomrule
\end{tabular}
\vskip -2mm
 \caption{Example of an inappropriately corrected error and an unchanged error in EFCamDat. We consider these types of errors to be dataset noise that might hinder GEC model performance.}
\vskip -5mm
\label{tab:noise}
\end{table}
For example, \citet{lo-etal-2018-cool} reported that an EncDec-based GEC model trained on EFCamDat~\citep{efcamdat}\footnote{\url{https://corpus.mml.cam.ac.uk/efcamdat2/public_html/}}, the largest publicly available learner corpus as of today (two million sentence pairs), was outperformed by a model trained on a smaller dataset (e.g., 720K pairs).
They hypothesized that this may be due to the noisiness of EFCamDat, i.e., the presence of sentence pairs whose correction still contained grammatical errors due to inappropriate edits or being left uncorrected. 
For example, in Table~\ref{tab:noise}, ``\textit{discuss about}'' should most likely have been corrected to ``\textit{discuss}'', and ``\textit{are discussing}'', respectively. 
We confirmed that there is a non-negligible amount of noise in commonly used GEC datasets (Section~\ref{sec:noise_gec}).

We recognise data noise as a generally overlooked issue in GEC, and consider the question of \textit{whether a better GEC model can be built by reducing noise in GEC corpora}.
To this end, we designed a \textit{self-refining} approach---an effective denoising method where residual errors left by careless or unskilled annotators are corrected by an existing GEC model.
This approach relies on the consistence of the GEC model's predictions (Section~\ref{sec:proposed_model}).

We evaluated the effectiveness of our method over several GEC datasets, and found that it considerably outperformed baseline methods, including three strong denoising baselines based on a filtering approach, which is a common approach in MT~\citep{bei-etal-2018-empirical,junczys-dowmunt2018wmt:filtering,rossenbach-etal-2018-rwth}.
We further improved the performance by applying task-specific techniques and achieved state-of-the-art performance on the CoNLL-2014, JFLEG, and BEA-2019 benchmarks.
Finally, through our analysis, we found unexpected benefits to our approach: (i) the approach benefits from the advantage of self-training in neural sequence generation due to its structural similarity (Section~\ref{subsec:self_training_benefit}), (ii) resulted in significant increase in recall while maintaining equal precision, indicating improved coverage of correction (Section~\ref{subsec:recall_improve}), and (iii) there seems to be a tendency for more fluent edits, possibly leading to more native-sounding corrections (Section~\ref{subsec:fluency_edits}). The last is reflected in performance on the JFLEG benchmark, which focuses on fluency edits.

In summary, we present a data denoising method which improves GEC performance, verify its effectiveness by comparing to both strong baselines and current best-performing models, and analyze how the method affects both GEC performance and the data itself.

\section{Related Work}
\label{sec:related_work}

In GEC, previous studies have generally focused on typical errors, such as the use of articles~\cite{Han:06:Journal}, prepositions~\cite{Felice:08:COLING}, and noun numbers \cite{Nagata:06:COLING}.
More recently, many studies have addressed GEC as a MT problem where ungrammatical text is expected to be {\it translated} into grammatical text.
This approach allows the adoption of sophisticated sequence-to-sequence architectures~\citep{sutskever:2014:NIPS,bahdanau:2015:ICLR, vaswani:2017:NIPS} that have achieved strong performance but require a large amount of data~\citep{chollampatt:2018:AAAI,junczys:2018:NAACL, kiyono-etal-2019-empirical}.
In GEC, the data are usually manually built by experts, which lead to an underlying assumption that the data is noise-free.
Therefore, to the best of our knowledge, noise in existing common datasets remains largely under-explored and no previous research has investigated the effectiveness of denoising GEC datasets.
Recently, \citet{lichtarge2020data} proposed a method for filtering large and noisy synthetic pretrained data in GEC by deriving example-level scores on their
pretrained data.\footnote{\citet{lichtarge2020data} has appeared after our submission.}
However, what they regard as noise consists of instances in source sentences (i.e., not target sentences) of the synthetic data that are outside the genuine learner error distribution, where they perform data selection based on the small and higher-quality genuine data (namely, the learner corpora we attempt to denoise in this study).
Therefore, our methods are not comparable, and it is expected to further improve the performance by combining both methods, which we plan to investigate in our future work.

In contrast, data noise is becoming an increasingly important topic in MT, where it is common to use automatically acquired parallel data via web crawling in addition to high-quality curated data.
As a result, the MT field faces various data quality issues such as misalignment and incorrect translations, which may significantly impact translation quality~\cite{khayrallah-koehn-2018-impact}.
A straightforward solution is to apply a filtering approach, where noisy data are filtered out and a smaller subset of high-quality sentence pairs is retained~\citep{bei-etal-2018-empirical,junczys-dowmunt2018wmt:filtering,rossenbach-etal-2018-rwth}.
Nevertheless, it is unclear whether such a filtering approach can be successfully applied to GEC, where commonly available datasets tend to be far smaller than those used in recent neural MT research.
Hence, in this study, we investigate its effectiveness by conducting a comparative experiment using the proposed denoising approach.

\section{Noise in GEC Datasets}
\label{sec:noise_gec}

\begin{table*}[!t]
\centering
\begin{tabular}{ll}
\toprule
(1) & BEA-train   \\ \midrule
$\bm{X}$ :   &  {\it  I will make a poet to kill this pain.}      \\
$\bm{Y}$ :  &  {\it I will make a poem to kill this pain.}      \\
$\bm{Y}^{\prime}$ :   & {\it I will write a poem to get rid of this pain.}                    \\
\midrule
(2)  &  EFCamDat \\  \midrule
$\bm{X}$ :   &  {\it The restaurant in front of movie teather.}      \\
$\bm{Y}$ :  &  {\it The restaurant in front of movie theater.}      \\
$\bm{Y}^{\prime}$ :   & {\it The restaurant is located opposite the movie theater.}                    \\
\midrule
(3)  & Lang-8 \\  \midrule
$\bm{X}$ :   &  {\it Coordinate with product support team for potential customer show site visit ;}      \\
$\bm{Y}$ :  &  {\it Coordinate with product support team for potential customer show site visits;}      \\
$\bm{Y}^{\prime}$ :   & {\it Please coordinate with the product support team to escort potential customers to site visits.}                    \\
\bottomrule
\end{tabular}
 \caption{Examples of original sources sentences ($\bm{X}$), original target sentences ($\bm{Y}$) and target sentences reviewed by the expert ($\bm{Y}^{\prime}$) in the most commonly used training data for GEC.}
\label{tab:noise_training}
\end{table*}

\begin{table}[t]
\centering
\begin{tabular}{lr}
\toprule
Dataset           & WER (\%noise) \\
\midrule
BEA-train               & 37.1   \\
EFCamDat          & 42.1    \\
Lang-8 & 34.6   \\ 
\bottomrule
\end{tabular}
\vskip -2mm
\caption{Amount of noise in GEC training data estimated by WER.}
\label{table:amount_of_noise}
\vskip -2mm
\end{table}

In this study, we define {\em noise} as two types of residual grammatical errors in target sentences: inappropriate edits and those left uncorrected (Table~\ref{tab:noise}).
Most learner corpora, such as EFCamDat and Lang-8~\citep{mizumoto:2011:IJCNLP,Tajiri:12:ACL}, are constructed based on correction logs in which the source texts are provided by human language learners and the corresponding corrected target texts are provided by editor (annotators).
Unless each annotator has 100\% accuracy, all corpora inevitably contain noise.

The presence of noise in GEC data was uncovered by previous work such as \citet{lo-etal-2018-cool}, but the exact nature of it was unexplored.
To confirm this, we manually assessed how much noise was contained in the following three commonly used training datasets: the BEA official training dataset (henceforth, \emph{BEA-train}) provided in the BEA-2019 workshop~\citep{bryant-etal-2019-bea}\footnote{See Appendix~\ref{appendix:bea-2019-dataset} for dataset details.}, EFCamDat, and the non-public Lang-8 corpus (henceforth, \emph{Lang-8})\footnote{A corpus consisting of correction logs from 2012 to 2019 in Lang-8. We recognize that BEA-train contains a subset of this corpus, but we use it without distinction in this study.}.
For 300 target sentences $\bm{Y}$ from each dataset, one expert reviewed them and we obtained denoised ones $\bm{Y}^{\prime}$ (Table~\ref{tab:noise_training}).
We then calculated the word edit rate (WER) between the original target sentences $\bm{Y}$ and the denoised target sentences $\bm{Y}^{\prime}$.
WER is defined as follows:
\begin{align}
\mbox{WER} = \frac{\sum_{i=1}^{N} d(\bm{Y}_i,\bm{Y}^{\prime}_i)}{\sum_{i=1}^{N} |\bm{Y}_i|}
\label{eq:wer}
\end{align}
where, $|\bm{Y}_i|$ is the total number of words in each original target sentences $\bm{Y}_i$ and $d(\cdot)$ is the word-based Levenshtein distance.
Table~\ref{table:amount_of_noise} shows the amount of noise in the datasets estimated by WER.
Here, the WER values are slightly higher than expected, but this is most likely caused by fluency edits by the editor, making the sentence more native-like.
Thus, we found that (i) there is a non-negligible amount of ``noise'' in the most commonly used training data for GEC, and (ii) EFCamDat is much noisier than the other two training datasets.

\section{Proposed Denoising Method}
\label{sec:proposed_model}
The supervised learning problem for GEC is formally defined as follows.
Let $\bm{\theta}$ be all trainable parameters of a GEC model, and $\mathcal{D}$ be training data consisting of pairs of an ungrammatical source sentence $\bm{X}$ and a grammatical target sentence $\bm{Y}$, i.e., $\mathcal{D}=\{(\bm{X}_i, \bm{Y}_i)\}_{i=1}^n$. 
Then, the objective is to find the optimal parameters $\widehat{\bm{\theta}}$ that minimize the following loss function $\mathcal{L}(\mathcal{D}, \bm{\theta})$ on training data $\mathcal{D}$:
\begin{align}
\mathcal{L}(\mathcal{D}, \bm{\theta}) &= - \frac{1}{\vert \mathcal{D} \vert} \sum_{(\bm{X}, \bm{Y}) \in \mathcal{D}}\!\!\!\log(p( \bm{Y} | \bm{X}, \bm{\theta})).
\label{eq:loss1}
\end{align}
Conventionally, training data $\mathcal{D}$ is assumed to be ``clean'' parallel data. 
However, as argued in Section~\ref{sec:noise_gec}, this assumption typically does not hold in GEC.
Here, we assume that training data $\mathcal{D}$ is ``noisy'', 
and, for clarity, we use the notation \truseteddata{} to represent ``clean'' parallel data, where ``clean'' means ``denoised'' in this context. 
The goal is, first, to obtain a new set \truseteddata{} by denoising $\mathcal{D}$, and then, to obtain a GEC model $\bm{\widehat{\theta}}$ on the new training data \truseteddata{}.

To deal with data noise, a straightforward solution is to apply a filtering approach, where noisy data are filtered out and a smaller subset of high-quality sentence pairs is retained, as employed in MT.
However, applying a filtering approach may not be the best choice in GEC for two reasons: (i) GEC is a low-resource task compared to MT, thus further reducing data size by filtering may be critically ineffective;
(ii) Even noisy instances may still be useful for training since they might contain some correct edits as well (Note that these correct edits would have also been lost to filtering, further decreasing the amount of informative cues in training).

\begin{algorithm}[t]
  \DontPrintSemicolon
  \KwData{Noisy Parallel Data \noisydata{}}
  \KwResult{Denoised Parallel Data \truseteddata{}}
   \truseteddata{} = \{\}\tcp*{create empty set}
   Train a base model and acquire $\bm{\theta}$ from \noisydata{} \;
  \For{$(\bm{X}, \bm{Y}) \in \noisydata{}$}{
   $\bm{Y}^{\prime}$ = $\mathrm{Beam\_Search\_Decoding}(\bm{Y}; \bm{\theta}$) \;
    Compute perplexity $\mathrm{PPL}(\bm{Y})$ and $\mathrm{PPL}(\bm{Y}^{\prime}$) \;
    \If{$\mathrm{PPL}(\bm{Y}) - \mathrm{PPL}(\bm{Y}^{\prime}) \geqq 0$}{
      $\bm{\hat{Y}} = {\bm{Y}^{\prime}}$\;
    }
     \Else{
     $\bm{\hat{Y}} = \bm{Y}$ \;
    }
    $\truseteddata{} =\truseteddata{} \cup \{(\bm{X}, \bm{\hat{Y}})\}$ \;
  }
    Train a denoised new model $\bm{\widehat{\theta}}$ from \truseteddata{} \;
    \caption{Denoising GEC parallel data with self-refinement}
    \label{alg:self-refinement}
\end{algorithm}

As an alternative to filtering, we propose a self-refinement (SR) approach for denoising GEC training data (Algorithm~\ref{alg:self-refinement}).
The main idea is to train a GEC model (henceforth, \emph{base model}) on noisy parallel data \noisydata{} and to use it for refining target sentences in \noisydata{}.
Noisy annotations are potentially caused by carelessness or insufficient skills of annotators.
This causes inconsistent corrections in similar context.
In contrast, machine learning-based GEC models, such as EncDec, tend to be reliably consistent given similar contexts.
Given noisy parallel data $\noisydata=\{(\bm{X}_i, \bm{Y}_i)\}_{i=1}^n$, we generate new target sentences $\bm{\hat{Y}}_i$ from the original target sentences $\bm{Y}_i$ and pair them with their original source sentences $\bm{X}_i$ (line 4 in Algorithm~\ref{alg:self-refinement}).
The consistency of the base model predictions ensures that the resulting parallel data $\truseteddata{}=\{(\bm{X}_i, \bm{\hat{Y}}_i)\}_{i=1}^n$ contain noise at a less extent.
It is worth noting that SR can be regarded as a variant of self-training due to its structural similarity, except that it takes the target sentences rather than the source sentences as input to the model. 
The algorithm itself is the key difference from existing methods based on self-training~\cite{wang-2019-revisiting,nie-etal-2019-simple,Xie_2020_CVPR}.

One challenge of this approach is that the base model may consistently make inaccurate corrections.
We thus incorporate a fail-safe mechanism as a sub-component to restore the original target sentence if the GEC model makes incorrect corrections (lines 5-9).
For example, in cases such as in Table~\ref{tab:noise}, the base model may predict every instance as ``discuss about". 
In this step, to determine whether to accept the output $\bm{Y}^{\prime}$ of the base model as a new target sentence, we compare the perplexity of the model output $\mathrm{PPL}(\bm{Y}^{\prime})$ with that of the original target sentence $\mathrm{PPL}(\bm{Y})$.
Language models are trained on native-written corpora, meaning they can reasonably be assumed to contain information needed to estimate grammaticality. 
We believe that a measure of perplexity is a straightforward approach to exploit this information.

\section{Experiments}
\label{sec:experiments}
We evaluate the proposed method in two ways.
First, we exclusively focus on investigating the effectiveness of the proposed denoising method (Section~\ref{subsec:results}). 
Then, we compare our strongest model trained with denoised data (henceforth, \textit{denoised model}), with current best-performing ones to investigate whether the proposed method has a complementary effect on existing task-specific techniques (Section~\ref{subsec:result_with_top}).

\subsection{Configurations}
\label{subsec:config}
\paragraph{Dataset}

For the training dataset, we used the same datasets as mentioned in Section~\ref{sec:noise_gec}: BEA-train, EFCamDat, and Lang-8.
In addition, we used the BEA official validation set (henceforth, \emph{BEA-valid}) provided in the BEA-2019 workshop as validation data.
The characteristics of the datasets are summarized in Table~\ref{table:data_summary}.
For preprocessing, we tokenized the training data using the \texttt{spaCy} tokenizer\footnote{\url{https://spacy.io/}}.
Then, we removed sentence pairs where both sentences where identical or both longer than 80 tokens.
Finally, we acquired subwords from the target sentence via the byte-pair-encoding (BPE)~\citep{sennrich:2016:ACL} algorithm.
We used the \texttt{subword-nmt} implementation\footnote{\url{https://github.com/rsennrich/subword-nmt}} and then applied BPE to splitting both source and target texts.
The number of merge operations was set to 8,000.

\begin{table}[t]
\centering
\small
\tabcolsep=1pt
\begin{tabular}{lrcc}
\toprule
Dataset          & \#sent (pairs)  & Split & Scorer  \\
\midrule
BEA-train        & 561,100 & train  & - \\
EFCamDAT         &  2,269,595  &   train & -    \\
Lang-8 &  5,689,213  &  train & -   \\
BEA-valid         & 2,377   &  valid  & -   \\
\midrule
CoNLL-2014       & 1,312   &   test & $ \rm M^2$~scorer \& GLEU  \\
JFLEG            & 747   &   test & $\rm M^2$~scorer \& GLEU \\
BEA-test         & 4,477   &  test      & ERRANT  \\
\bottomrule
\end{tabular}
\vskip -2mm
\caption{Summary of datasets used in our experiments.}
\label{table:data_summary}
\end{table}

\paragraph{Evaluation}
To investigate the effectiveness of the proposed method, we followed the work by \citet{mita-etal-2019-cross} and evaluated the performance of the GEC models across various GEC datasets in terms of the same evaluation metrics.
We report the results measured by both $ \rm M^2$~scorer~\citep{dahlmeier:2012:M2}\footnote{\url{https://github.com/nusnlp/m2scorer/releases}} and GLEU metric~\citep{napoles:2015:ACL,napoles:2016:gleu}\footnote{\url{https://github.com/cnap/gec-ranking}} on both the CoNLL-2014 test set and the JFLEG test set~\citep{Napoles:17:EACL}.
All reported results (except those corresponding to the ensemble models) are the average of three distinct trials using three different random seeds.
Let us emphasize that our focus is on denoising the \emph{training} data, and denoising the test data is out of the scope of this study.
The commonly used test data, such as CoNLL-2014 and JFLEG, have multiple references which can lower the noise factor.
In addition to having multiple references, both JFLEG and CoNLL-2014 have been specifically constructed for GEC evaluation, while the training data (Lang-8 and EFCamDat) are more of an organic collection of learner and editor interactions. 
Naturally, we believe it is reasonable to assume that the test data are considerably cleaner.

\paragraph{Model}
We employed the ``Transformer (big)'' settings~\citet{vaswani:2017:NIPS} using the implementation in the \texttt{fairseq} toolkit~\citep{ott2019:arxiv:fairseq}.
Details on the hyper-parameters are listed in Appendix~\ref{appendix:hyper-parameter-settings}.
As a language model for the fail-safe mechanism, we used the PyTorch implementation of GPT-2~\citep{radford2019language}\footnote{\url{https://github.com/huggingface/transformers}}.
Note that to avoid a preference for shorter phrases, we normalized the perplexity by sentence length.

\begin{figure*}[t]
    \begin{tabular}{c}
    
  \begin{minipage}[]{0.45\linewidth}
    \centering
    \includegraphics[width=1.0\linewidth]{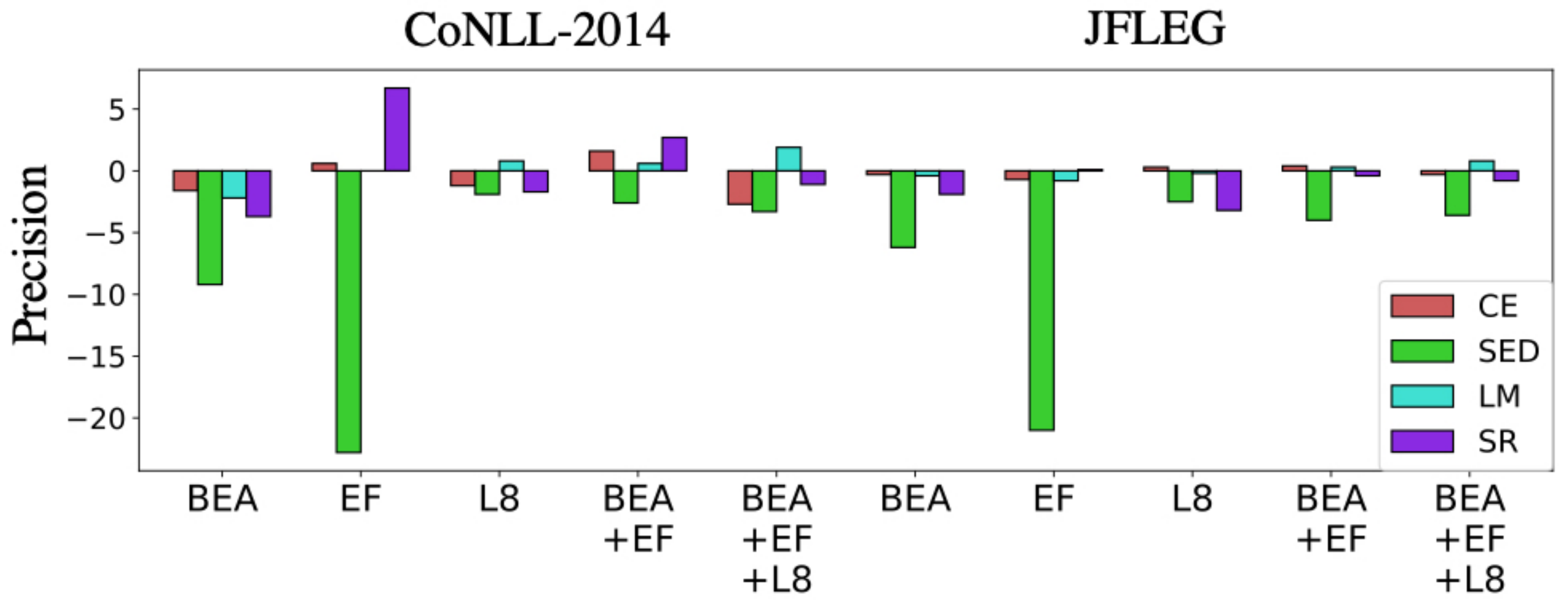}    
    \subcaption{Precision}\label{fig:main_prec}
  \end{minipage}

    \begin{minipage}{0.06\hsize}
        \hspace{1mm}
    \end{minipage}
    \begin{minipage}[]{0.45\linewidth}
    \centering
    \includegraphics[width=1.0\linewidth]{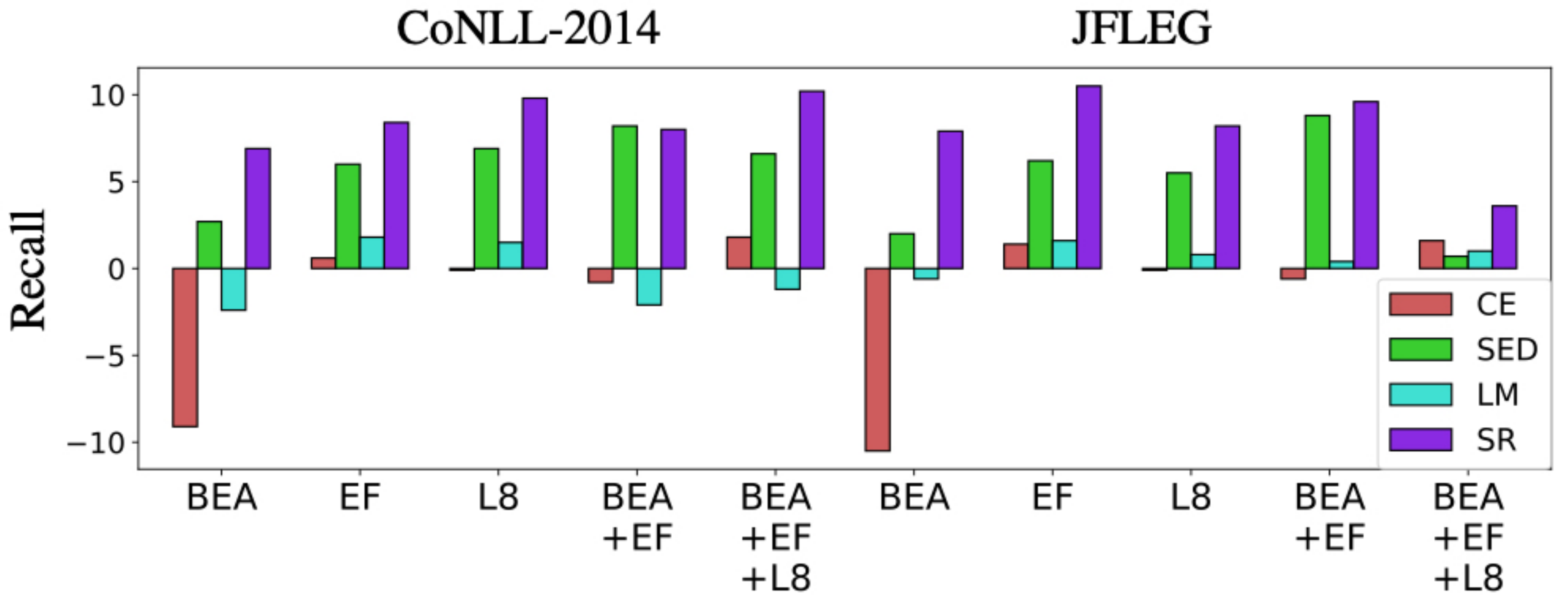}    
    \subcaption{Recall}\label{fig:main_rec}
  \end{minipage}

  \end{tabular}
  \caption{Increases and decreases in precision and recall for denoising methods when no denoising is set to 0.}
  \label{fig:main_prec_rec}
\end{figure*}

\subsection{Baselines}
As argued in Section~\ref{sec:proposed_model}, we hypothesized that the filtering-based denoising approaches are not well-suited for GEC.
To verify this hypothesis, we employed the following three filtering-based denoising baseline methods in addition to a base model trained in noisy parallel data \noisydata{} (henceforth, \emph{no denoising}).

\paragraph{Cross-entropy filtering (CE filtering)}
The dual conditional cross-entropy filtering method was proposed by ~\citet{junczys-dowmunt2018wmt:filtering} and achieved the highest performance on the noisy parallel corpus filtering task at WMT2018~\citep{koehn-etal-2018-findings}.
In this study, we prepared forward and reverse pre-train models using the BEA-train dataset to adapt the filtering method to GEC. 
We obtained the filtered data by removing 20\% of the sentence pairs\footnote{The value is derived from a preliminary experiment (Appendix~\ref{appendix:cross-entorpy_exp}).} with higher scores from the training data and used them for training.

\paragraph{Sentence-level error detection filtering (SED filtering)}
\citet{asano:2019:bea} demonstrated the effectiveness of the sentence-level error detection (SED) model as a filtering tool to preprocess GEC input.
Considering these findings, we adopted SED as a filtering-based denoising method for training data. 
More specifically, we discarded the source-target sentence pairs in the noisy parallel data~\noisydata{} if the SED model predicted the target sentence as an incorrect one.
Following~\citet{asano:2019:bea}, we obtained binary-labeled data using the BEA-train dataset to prepare a training set for the SED model, and then fine-tuned BERT~\citep{devlin-etal-2019-bert} on the prepared data.

\paragraph{Language model filtering (LM filtering)}
Language model-based filtering is a method based on the hypothesis that if the perplexity of a target sentence is larger than that of the source sentence, the target sentence is more likely to contain noise.
LM filtering has the same motivation as the one underlying the fail-safe mechanism.
We used GPT-2 as the pre-trained language model.

\begin{table}[t!]
\centering
\small
\tabcolsep=3pt
\scalebox{1.0}[1.0]{ 
\begin{tabular}{lcccc}
\toprule
  & \multicolumn{2}{c}{\begin{tabular}{c} CoNLL-2014 \end{tabular}} & \multicolumn{2}{c}{\begin{tabular}{c} JFLEG \end{tabular}}  \\
\cmidrule(r){2-3}\cmidrule(r){4-5}
\multicolumn{1}{l}{Model}  & \fscore{}  & GLEU   & \fscore{}  & GLEU  \\
\midrule
\scriptsize{\textbf{BEA-train (BEA):}} \\
No denoising $_{\textbf{BEA}}$      & 49.6 & 63.3    & 58.7  & 52.3    \\
CE filtering $_{\textbf{BEA}}$       & 42.9 & 61.0    & 52.7  & 49.0      \\
SED filtering $_{\textbf{BEA}}$      & 45.1 & 62.7    & 55.6       &  52.2      \\
LM filtering $_{\textbf{BEA}}$       & 47.1 & 63.0    & 58.3       &  52.6       \\
SR $_{\textbf{BEA}}$ (\textbf{Ours}) & \textbf{50.3}  & \textbf{64.2} & \textbf{60.5}  & \textbf{54.8}   \\ \midrule
\scriptsize{\textbf{EFCamDAT (EF):}} \\
No denoising $_{\textbf{EF}}$       & 40.3 & 61.3    & 59.5  & 53.7    \\
CE filtering $_{\textbf{EF}}$        & 40.9 & 61.5    &  59.8  & 54.2      \\
SED filtering $_{\textbf{EF}}$       & 26.5  & 54.0   & 47.5  & 49.7   \\
LM filtering $_{\textbf{EF}}$         & 41.2  & 61.7  & 59.7   & 54.2       \\
SR $_{\textbf{EF}}$ (\textbf{Ours}) & \textbf{48.4}  & \textbf{63.5}   & \textbf{63.9}  & \textbf{57.1}   \\ \midrule
\scriptsize{\textbf{Lang-8 (L8):}} \\
No denoising $_{\textbf{L8}}$       & 54.9 & 65.9     & 68.4 & 58.1    \\
CE filtering $_{\textbf{L8}}$        & 54.1 & 65.3     &\bf 68.6 &  58.2     \\
SED filtering $_{\textbf{L8}}$       & 55.7 & 67.1     & 68.5 & 60.7 \\
LM filtering $_{\textbf{L8}}$        & 55.9 & 66.3     & \textbf{68.6}       &  59.1       \\
SR $_{\textbf{L8}}$ (\textbf{Ours})  &\textbf{56.5} &\textbf{67.7} &  \textbf{68.6} &  \textbf{61.0}  \\ \midrule
\scriptsize{\textbf{BEA+EF:}} \\
No denoising $_{\textbf{BEA}+\textbf{EF}}$   & 49.1  & 63.4       & 62.0  & 53.9  \\
CE filtering $_{\textbf{BEA}+\textbf{EF}}$    & 49.6  & 63.3       & 61.9   & 54.5      \\
SED filtering  $_{\textbf{BEA}+\textbf{EF}}$  & 51.2  & 64.9       & 62.8  & 56.7   \\
LM filtering$_{\textbf{BEA}+\textbf{EF}}$     & 48.3  & 63.3       & 62.3       &  54.7     \\
SR $_{\textbf{BEA}+\textbf{EF}}$ (\textbf{Ours})  & \textbf{54.5}& \textbf{65.2}  & \textbf{65.5} & \textbf{58.0} \\ \midrule
\scriptsize{\textbf{BEA+EF+L8:}} \\
No denoising $_{\textbf{BEA}+\textbf{EF}+\textbf{L8}}$  & 56.1  & 65.7    & 67.0 & 56.9   \\
CE filtering $_{\textbf{BEA}+\textbf{EF}+\textbf{L8}}$   & 55.0  & 66.0    & 68.6 & 58.2      \\
SED filtering $_{\textbf{BEA}+\textbf{EF}+\textbf{L8}}$  & 56.1  & 67.3    & 67.7 & 60.3    \\
LM filtering $_{\textbf{BEA}+\textbf{EF}+\textbf{L8}}$   & 56.7  & 65.9    & 68.0 &  57.8        \\
SR $_{\textbf{BEA}+\textbf{EF}+\textbf{L8}}$ (\textbf{Ours})  &  \textbf{58.8} &  \textbf{68.0} &  \textbf{70.6} &  \textbf{61.4}  \\
\bottomrule
\end{tabular} 
}
\vskip -2mm
\caption{Result of denoising experiments with cross-corpora evaluation: a \textbf{bold} value indicates the best result in each training data.}
 \label{tab:main_result}
\end{table}

\begin{table*}[t]
\centering
\begin{tabular}{lrrr} \toprule
Filtering method & \multicolumn{1}{c}{BEA-train} & \multicolumn{1}{c}{EFCamDat} & \multicolumn{1}{c}{Lang-8} \\ \midrule
CE filtering & 448,880 ($\triangledown$ 20.0\%) & 1,815,676  ($\triangledown$ 20.0\%)              & 4,551,370 ($\triangledown$ 20.0\%)          \\
SED filtering     & 317,957 ($\triangledown$ 43.3\%) & 1,250,744  ($\triangledown$ 44.9\%)              & 3,314,440 ($\triangledown$ 41.7\%)          \\
LM filtering      & 456,347  ($\triangledown$ 18.7\%)         & 1,936,238    ($\triangledown$ 14.7\%)         & 4,651,085  ($\triangledown$ 18.2\%)
\\ \bottomrule
\end{tabular}
\vskip -2mm
    \caption{The size of the filtered data. The numbers in parentheses indicate each reduction rates.}
    \label{tab:filterd_data}
\end{table*}

\subsection{Results}
\label{subsec:results}

Table~\ref{tab:main_result} shows the results of the main experiment.
The experimental results show that SR significantly outperformed the others, including the three strong denoising baseline models on the multiple datasets.
Applying SR to EFCamDat, for instance, yielded a larger performance improvement than without denoising (e.g, no denoising $_{\textbf{EF}}=40.3$ vs SR $_{\textbf{EF}}=48.4$ in CoNLL-2014 when using~\fscore{}).
Notably, we observed a similar trend when using both BEA-train and Lang-8 datasets as the training data, which indicated that SR was potentially effective for any corpora, not being limited to EFCamDat.

Furthermore, we compared the effectiveness of SR to other denoising methods.
The filtering-based methods, such as SED and LM filtering, generally achieved better results compared to the baseline models; however, they resulted in lower performance in smaller datasets such as BEA.
This could be caused by the fact that these filtering methods have filtered out the training instances containing not only noise but also many correct corrections that may still be partially useful for training.
As shown in Table~\ref{tab:filterd_data}, we analyzed the size of each training dataset after filtering.

Figure~\ref{fig:main_prec_rec} shows the increases and decreases in precision and recall when the performance without denoising is set as 0.
The experimental results show that there was a certain pattern underlying the denoising effect.
More specifically, reducing the noise by SR has little impact on the precision, but it has significantly improved the recall, indicating improved coverage of correction.
We provide the detailed analysis on this question in Section~\ref{subsec:recall_improve}.

\subsection{Comparison with Existing Models}
\label{subsec:result_with_top}

\begin{table*}[t!]
\centering
\small
\begin{tabular}{lccccc}
\toprule
  & \multicolumn{2}{c}{\begin{tabular}{c} CoNLL-2014 \end{tabular}} & \multicolumn{2}{c}{\begin{tabular}{c} JFLEG \end{tabular}}& \multicolumn{1}{c}{\begin{tabular}{c} BEA  \end{tabular}} \\
\cmidrule(r){2-3}\cmidrule(r){4-5}\cmidrule(r){6-6}
\multicolumn{1}{c}{Model}       & \fscore{}  & GLEU    & \fscore{}  & GLEU  & \fscore{} \\
\midrule
{\textbf{Single model:}}\\
 \citet{junczys:2018:NAACL}            & 53.0   & -       & -    & 57.9 & -  \\
 \citet{lichtarge-etal-2019-corpora}   & 56.8   & -       & -    & 61.6 & -  \\
 \citet{awasthi-etal-2019-parallel}    & 59.7   & -       & -    & 60.3 & -  \\
 \citet{kiyono-etal-2019-empirical}    & 61.3   & 68.6    & 71.3 & 59.7 & 64.2 \\
 SR +\textsc{PRET}+\textsc{SED}   
 & \textbf{61.4}   & \textbf{69.3} & \textbf{72.5} & \textbf{63.3} & \textbf{65.5}   \\ 
\midrule
{\textbf{Ensemble model:}} \\
 \citet{junczys:2018:NAACL}              & 55.8   & -       & -    & 59.9 & -  \\
     \citet{lichtarge-etal-2019-corpora} & 60.4   & -       & -    & 63.3 & -  \\
 \citet{grundkiewicz:2019:bea}           & 64.2   & -       & -    & 61.2 & 69.5 \\
  \citet{kiyono-etal-2019-empirical}     & \textbf{65.0}   & 68.8    & 72.9 & 61.4 & \textbf{70.2} \\
 SR +\textsc{PRET}+\textsc{R2L}+\textsc{SED}
                                         & 63.1  &\textbf{69.8}  & \textbf{73.9}  &\bf 63.7  & 67.8  \\ 

\bottomrule
\end{tabular}
\vskip -2mm
\caption{Comparison with existing top models: a \textbf{bold} value denotes the best result within the column. Both SR and BEA indicate SR $_{\textbf{BEA}+\textbf{EF}+\textbf{L8}}$ and BEA-test, respectively.}
\label{tab:exp_top}
\end{table*}

In the second experiment, we compared our best denoised model with the current best performing models to investigate whether SR works well with existing task-specific techniques. 
We incorporated task-specific techniques that have been widely used in shared tasks such as BEA-2019 and WMT-2019\footnote{\url{http://www.statmt.org/wmt19/}} into the proposed denoised model to further improve the performance. 
Concerning the task-specific techniques, we followed the work reported by~\citet{kiyono-etal-2019-empirical}, as detailed below.

\paragraph{Pre-training with pseudo data} (\textsc{PRET})\hspace*{3mm}
\citet{kiyono-etal-2019-empirical} investigated the applicability of incorporating pseudo data into the model and confirmed the reliability of their proposed settings by showing acceptable performance on several datasets.
We trained the proposed model using their pre-trained model ``\textsc{PretLarge}+\textsc{SSE}'' settings\footnote{\url{https://github.com/butsugiri/gec-pseudodata}}.

\paragraph{Right-to-left re-ranking} (\textsc{R2L})\hspace*{3mm} 
R2L is a common approach used to improve model performance by re-ranking using right-to-left models trained in the reverse direction~\citep{sennrich:2016:wmt, sennrich:2017:wmt} in MT.
More recently, previous studies confirmed the effectiveness of this approach when applied to GEC~\citep{ge2018reaching, grundkiewicz:2019:bea}.
We adapted \textsc{R2L} to the proposed model. 
Specifically, we generated $n$-best hypotheses using an ensemble of four left-to-right (L2R) models and then re-scored the hypotheses using these models.
We then re-ranked the $n$-best hypotheses based on the sum of the both two scores.

\paragraph{Sentence-level error detection} (\textsc{SED})\hspace*{3mm}
\textsc{SED} is used to identify whether a given sentence contains any grammatical errors.
Following the work presented by~\citet{asano:2019:bea}, we employed a strategy based on reducing the number of false positives by only considering sentences that contained grammatical errors in the GEC model, using an \textsc{SED} model.
We implemented the same model employed for~\textsc{SED} filtering.

\vspace{5mm}
We evaluated the performance of the proposed best denoised model incorporated with the task-specific techniques on the three existing benchmarks: CoNLL-2014, JFLEG, and BEA-test, and then compared the scores with existing best-performing models.
Table~\ref{tab:exp_top} shows the results for both the single and the ensemble models after applying \textsc{PRET}, \textsc{SED}\footnote{See Appendix~\ref{appendix:ablation_sed} for an ablation study of SED.}, and \textsc{R2L} to SR\footnote{Improved results on the CoNLL-2014 and BEA-2019 have been appeared in arXiv less than 3 months before our submission~\citep{Kaneko2020EncoderDecoderMC,omelianchuk2020gector} that are considered contemporaneous to our submission. More detailed experimental results, including a comparison with them, are presented in Appendix~\ref{appendix:top_model} for reference.}.
Since the reference of BEA-test is publicly unavailable, we evaluated the models on \texttt{CodaLab}\footnote{\url{https://competitions.codalab.org}} under the rules of BEA-2019 workshop.
We confirmed that our best denoised model works complementarily with existing task-specific techniques, as compared with the performance presented in Table~\ref{tab:main_result}.
As a result, our best denoised model achieved state-of-the-art performance on the CoNLL-2014, JFLEG, and BEA-2019 benchmarks.
Noteworthy is that the proposed model achieved state-of-the-art results on the JFLEG benchmark in terms of both single ($\text{GLEU}=63.3$) and ensemble results ($\text{GLEU}=63.7$). 
We provide a detailed analysis on this question in Section~\ref{subsec:fluency_edits}.

\begin{table}[t!]
    \centering
     \small
    \begin{tabular}{ll}
    \toprule 
        \rownumber
        & \textbf{Improved by denoising (66.4\%)}\\ \midrule
        \multicolumn{2}{l}{
            \tabcolsep 2pt
            \begin{tabular}{rp{14.5em}}
                \rx & how about to going to movie . \\
                \ry & How about to going to movie . \\
                \ryy & How about going to a movie . \\
            \end{tabular}
        } \\

          \midrule 
        \rownumber 
        & \textbf{Both are correct (7.2 \%)}\\ \midrule
        \multicolumn{2}{l}{
            \tabcolsep 2pt
            \begin{tabular}{rp{14.5em}}
                \rx & I'm twenty-nine old. \\
                \ry & I'm twenty-nine years old. \\
                \ryy & I'm 29 years old. \\
            \end{tabular}
        } \\

         \midrule
        \rownumber
        & \textbf{Meaning is not preserved (10.4 \%)}\\ \midrule
        \multicolumn{2}{l}{
            \tabcolsep 2pt
            \begin{tabular}{rp{14.5em}}
                \rx & you need keep calm. \\
                \ry & You need to keep calm. \\
                \ryy & You need to be calm. \\
            \end{tabular}
        } \\
        
        \midrule 
        \rownumber 
        & \textbf{Added Unnecessary information (8.8 \%)} \\ \midrule
        \multicolumn{2}{l}{
            \tabcolsep 2pt
            \begin{tabular}{rp{14.5em}}
                \rx & The are a few of chair and desk. \\
                \ry & There are a few chairs and desks. \\
                \ryy & There are a few chairs and desks too. \\
            \end{tabular}
        } \\

        \midrule 
        \rownumber
        & \textbf{Contains errors (3.8 \%)} \\ \midrule
        \multicolumn{2}{l}{
            \tabcolsep 2pt
            \begin{tabular}{rp{14.5em}}
                \rx & There are very positive news for us. \\
                \ry & There is very positive news for us . \\
                \ryy & There is a very positive news for us . \\
            \end{tabular}
        } \\

        \midrule
        \rownumber
        & \textbf{Lack of fluency (3.4 \%)} \\ \midrule
        \multicolumn{2}{l}{
            \tabcolsep 2pt
            \begin{tabular}{rp{14.5em}}
                \rx & I go in my work on the bike. \\
                \ry & I go to work by bike. \\
                \ryy & I go to work on my bike. \\
            \end{tabular}
        } \\

        \bottomrule
    \end{tabular}
    \vskip -2mm
    \caption{Result of manual evaluation. Samples of input sentences ($\bm{X}$), original target sentences ($\bm{Y}$) and generated target sentences by our methods ($\bm{Y}^{\prime}$).}
    \label{tab:example_manual}
\end{table}

\begin{table*}[t!]
\centering
 \small
\begin{tabular}{lccccccccc}
\toprule
  & \multicolumn{3}{c}{\begin{tabular}{c} EF \end{tabular}} & \multicolumn{3}{c}{\begin{tabular}{c}BEA+EF  \end{tabular}}& \multicolumn{3}{c}{\begin{tabular}{c} BEA+EF+Lang-8 \end{tabular}} \\
\cmidrule(r){2-4}\cmidrule(r){5-7}\cmidrule(r){8-10}
\multicolumn{1}{c}{Model}     & Prec.     & Rec.     & \fscore{}    & Prec. & Rec.     & \fscore{} & Prec. & Rec. & \fscore{} \\
\midrule
 No denoising            & 48.5 & 24.0 & 40.3 & 58.5  & 30.0 & 49.1 & 62.8 & 39.2 & 56.1 \\
 SR w/o fail-safe & 49.8 & 32.3 & 44.3 & 57.4  & \textbf{41.0} & 53.1 & 59.5 & 45.8 & 56.1   \\ 
  SR             & \textbf{55.2} & \textbf{32.4} & \textbf{48.4} & \textbf{61.2}  & 38.0 & \textbf{54.5} & \textbf{61.7} & \textbf{49.4} & \textbf{58.8} \\
\bottomrule
\end{tabular}
\vskip -2mm
\caption{Ablation study of the fail-safe mechanism.}
\label{tab:ablation}
\end{table*}

\begin{table}[t!]
    \centering
    \small
     \tabcolsep 1pt
    \begin{tabular}{lll}
        \toprule 
        \multicolumn{2}{l}{\textbf{1 : Fail-safe deactivates:}} & \texttt{ppl.}\\
 \cmidrule(r){3-3}
                \rx & By the way, I have to *discuss of the education.&{94.65} \\
                \ry & By the way, I have to *discuss about education. &{79.64} \\ 
                \ryy & By the way, I have to discuss education.       &{73.37} \\
         \midrule
        \multicolumn{2}{l}{\textbf{2 : Fail-safe activates:}} & \texttt{ppl.}\\
 \cmidrule(r){3-3}
                \rx & Then I was treated in the hospital for one month. &{34.34} \\ 
                \ry & I was treated in the hospital for one month.      &{32.42} \\
                \ryy & I was treated *at the hospital for one month.    &{33.59} \\
        \bottomrule
\end{tabular}
    \vskip -2mm
    \caption{Examples of input sentences ($\bm{X}$), original target sentences ($\bm{Y}$) and generated target sentences by our methods ($\bm{Y}^{\prime}$) when our method activates and deactivates the fail-safe in EFCamDat. \texttt{ppl.} indicates perplexity.}
    \label{tab:activate_failsafe}
\end{table}

\section{Analysis}
\label{sec:analysis}

\subsection{Noise Reduction}
\label{subsec:noise-reduction}

To evaluate the quality of the dataset after denoising, a researcher with a high level of English proficiency (not involved with this work) manually evaluated 500 triples of source sentences $\bm{X}$, original target sentences $\bm{Y}$, and generated target sentences $\bm{Y}^{\prime}$ obtained by applying SR to EFCamDat satisfying $\bm{X} \neq \bm{Y} \neq \bm{\hat{Y}}$ (Table~\ref{tab:example_manual}).
We can see that 73.6\% of the replaced samples were determined to be appropriate corrections, including cases where both were correct. 
For reference, we provide examples of a confusion set before and after denoising in the Appendix~\ref{appendix:confusion_set}.

\subsection{Effect of the Fail-safe Mechanism}
\label{subsec:fail-safe-effect}
Next, we quantitatively and qualitatively analyzed the effectiveness of the fail-safe mechanism integrated into SR. 

Quantitatively, Table~\ref{tab:ablation} provides the results of the ablation study of the fail-safe mechanism on CoNLL-2014.
Our main proposal was to include a self-refining step to clean up training data, but we found that the added fail-safe mechanism serves as a sub-component to further improve performance.

Qualitatively, we directly observed the decisions of the fail-safe mechanism and how it affected denoising.
Table~\ref{tab:activate_failsafe} provides examples for cases when SR activates and deactivates the fail-safe mechanism in EFCamDat.
In the upper example (Table~\ref{tab:activate_failsafe}-1), \emph{*discuss of} in the source sentence should have been corrected to \emph{discuss}; however, it was inaccurately edited to~\emph{*discuss about} in the target sentence.
In this case, SR succeeded in selecting the correct model output with a lower perplexity without activating the fail-safe mechanism.
On the other hand, in the lower example, the model made an incorrect ``correction'' (\emph{*in $\rightarrow$ at}).
However, SR successfully activated the fail-safe mechanism and thus retained the correct original target sentence.

\vspace{5mm}
\subsection{Benefits from Self-training}
\label{subsec:self_training_benefit}
\begin{table}[t]
\centering
\begin{tabular}{lrr}
\toprule
Model           & \fscore{} & GLEU \\
\midrule
No denoising $_{\textbf{BEA}}$    &   49.6   &   63.3   \\
SR $_{\textbf{BEA}}$ w/ dropout  &   \textbf{50.3}   &   \textbf{64.2}   \\
SR $_{\textbf{BEA}}$ w/o dropout &    49.5  &    63.9  \\
\midrule
No denoising $_{\textbf{EF}}$     &   40.3   &    61.3  \\
SR $_{\textbf{EF}}$ w/ dropout   &   \textbf{48.4}   &    \textbf{63.5}  \\
SR $_{\textbf{EF}}$ w/o dropout  &     47.3 &   63.0 \\
\midrule
No denoising $_{\textbf{L8}}$     &   54.9   &    65.9  \\
SR $_{\textbf{L8}}$ w/ dropout   &   \textbf{56.5}   &    \textbf{67.7}  \\
SR $_{\textbf{L8}}$ w/o dropout  &     55.6 &   67.5 \\

\bottomrule
\end{tabular}
\vskip -2mm
\caption{Ablation study on the influence of dropout.}
\label{tab:dropout}
\end{table}

SR performed surprisingly well considering its simplicity.
One reason might be that SR benefited from the advantages of self-training, as it could be regarded as a variant of self-training (Section~\ref{sec:proposed_model}).
\citet{He2020Revisiting} investigated the effect of self-training in neural sequence generation and found that the dropout in the pseudo-training step (namely, the training step of the denoised model in this study) played an important role in providing a smoothing effect, meaning that semantically similar inputs were mapped to the same or similar targets.
As GEC also ideally holds the assumption of generating consistent targets for a similar context, this smoothing effect could contribute to avoiding over-fitting and improving fitting the target distribution in the pseudo-training step.
In fact, we confirmed that performance deteriorated when dropout was not applied in the training step of the denoised model, as shown in Table~\ref{tab:dropout}.
In the case of relatively noisy data such as EFCamDat and Lang-8, the performance was better than without denoising, even without dropout. 
This could be explained by the presence of the denoising effect that was the objective of this study.

\subsection{On the Increase of Recall}
\label{subsec:recall_improve}
A pattern emerged when denoising with SR---recall significantly increased, while precision was mostly maintained (Figure~\ref{fig:main_prec_rec}).
To clarify this observation, we manually assessed the amount of noise before and after denoising.
Specifically, in the same way as in Section~\ref{sec:noise_gec}, we asked the expert to review 500 samples of the target sentence before denoising $\bm{Y}$ and the target sentences after denoising $\bm{\hat{Y}}$. We then calculated the amount of noise using WER (Eq.\ref{eq:wer}).
As a result, we observed a decrease in WER from 43.2\% to 31.3\% before and after denoising, respectively.
This can be interpreted as (i) a large part of the noise was due to uncorrected errors, and (ii) the effect on model training was to correct the bias towards leaving errors unedited, resulting in higher recall.

\subsection{Facilitating Fluency Edits}
The results presented in Table~\ref{tab:exp_top} indicate that the proposed denoised model tends to (i) perform better on JFLEG and (ii) be specifically highly rated in GLEU compared to other best-performing models.
JFLEG was proposed by~\citet{Napoles:17:EACL} for the development and evaluation of GEC models in terms of fluency and grammaticality, i.e., making a sentence more native sounding.
Moreover, they showed that GLEU was correlated more strongly with humans than $ \rm M^2$ in JFLEG.
The fact that SR is rated higher on JFLEG using GLEU than other best-performing models can be interpreted as achieving more fluent editing.
One reason might be that SR performs a perplexity check on both the original target sentences and the new ones obtained after denoising, which always results in $\mathrm{PPL}(\bm{Y}) \geqq \mathrm{PPL}(\bm{\hat{Y}})$ between~\noisydata{} and~\truseteddata{}. 
Therefore, SR can be expected to refine not only grammaticality but also fluency of the target sentences, and as a result, the proposed denoised model is capable of performing more native-sounding corrections.
\label{subsec:fluency_edits}

\section{Conclusion and Future Work}
In this study, we focused on the quality of GEC datasets.
The motivation behind our study was based on the hypothesis that the carelessness or insufficient skill of the annotators involved in data annotation could often lead to producing noisy datasets.
To address this problem, we presented a self-refinement approach as a simple but effective denoising method which improved GEC performance, and verified its effectiveness by comparing to both strong baselines based on filtering approach and current best-performing models.
Furthermore, we analyzed how SR affects both GEC performance and the data itself.

Recently, several methods that incorporate pre-trained masked language models such as BERT, XLNet~\citep{xlnet-nips2019}, and RoBERTa~\citep{Liu2019RoBERTaAR} into EncDec based GEC have been proposed and achieved remarkable results~\citep{Kaneko2020EncoderDecoderMC,omelianchuk2020gector}.
These approaches modify the model architecture and do not directly compete with the data-driven approaches discussed in this study. 
Thus, the combination of these methods can be expected to further improve the performance, which we plan to investigate in our future work.

\section*{Acknowledgments}
We thank the Tohoku NLP laboratory members who provided us with their valuable advice. 
We are grateful to Tomoya Mizumoto and Ana Brassard for their insightful comments and suggestions.

\bibliography{reference}

\begin{thebibliography}{54}
\expandafter\ifx\csname natexlab\endcsname\relax\def\natexlab#1{#1}\fi

\bibitem[{Asano et~al.(2019)Asano, Mita, Mizumoto, and Suzuki}]{asano:2019:bea}
Hiroki Asano, Masato Mita, Tomoya Mizumoto, and Jun Suzuki. 2019.
\newblock {The {AIP}-Tohoku System at the {BEA}-2019 Shared Task}.
\newblock In \emph{Proceedings of the Fourteenth Workshop on Innovative Use of
  NLP for Building Educational Applications (BEA 2019)}, pages 176--182.

\bibitem[{Awasthi et~al.(2019)Awasthi, Sarawagi, Goyal, Ghosh, and
  Piratla}]{awasthi-etal-2019-parallel}
Abhijeet Awasthi, Sunita Sarawagi, Rasna Goyal, Sabyasachi Ghosh, and Vihari
  Piratla. 2019.
\newblock {Parallel Iterative Edit Models for Local Sequence Transduction}.
\newblock In \emph{Proceedings of the 2019 Conference on Empirical Methods in
  Natural Language Processing and the 9th International Joint Conference on
  Natural Language Processing (EMNLP-IJCNLP 2019)}, pages 4259--4269.

\bibitem[{Bahdanau et~al.(2015)Bahdanau, Cho, and Bengio}]{bahdanau:2015:ICLR}
Dzmitry Bahdanau, Kyunghyun Cho, and Yoshua Bengio. 2015.
\newblock {Neural Machine Translation by Jointly Learning to Align and
  Translate}.
\newblock In \emph{Proceedings of the 3rd International Conference on Learning
  Representations (ICLR 2015)}.

\bibitem[{Bei et~al.(2018)Bei, Zong, Wang, Fan, Li, and
  Yuan}]{bei-etal-2018-empirical}
Chao Bei, Hao Zong, Yiming Wang, Baoyong Fan, Shiqi Li, and Conghu Yuan. 2018.
\newblock {An Empirical Study of Machine Translation for the Shared Task of
  {WMT}18}.
\newblock In \emph{Proceedings of the Third Conference on Machine Translation
  (WMT 2018): Shared Task Papers}, pages 340--344.

\bibitem[{Brockett et~al.(2006)Brockett, Dolan, and
  Gamon}]{brockett-etal-2006-correcting}
Chris Brockett, William~B. Dolan, and Michael Gamon. 2006.
\newblock {Correcting {ESL} Errors Using Phrasal {SMT} Techniques}.
\newblock In \emph{Proceedings of the 21st International Conference on
  Computational Linguistics and 44th Annual Meeting of the Association for
  Computational Linguistics (COLING-ACL 2006)}, pages 249--256.

\bibitem[{Bryant et~al.(2019)Bryant, Felice, Andersen, and
  Briscoe}]{bryant-etal-2019-bea}
Christopher Bryant, Mariano Felice, {\O}istein~E. Andersen, and Ted Briscoe.
  2019.
\newblock The {BEA}-2019 shared task on grammatical error correction.
\newblock In \emph{Proceedings of the Fourteenth Workshop on Innovative Use of
  NLP for Building Educational Applications (BEA 2019)}, pages 52--75.

\bibitem[{Chollampatt and Ng(2018)}]{chollampatt:2018:AAAI}
Shamil Chollampatt and Hwee~Tou Ng. 2018.
\newblock {A Multilayer Convolutional Encoder-Decoder Neural Network for
  Grammatical Error Correction}.
\newblock In \emph{Proceedings of the Thirty-Second AAAI Conference on
  Artificial Intelligence (AAAI 2018)}, pages 5755--5762.

\bibitem[{Dahlmeier and Ng(2012)}]{dahlmeier:2012:M2}
Daniel Dahlmeier and Hwee~Tou Ng. 2012.
\newblock {Better Evaluation for Grammatical Error Correction}.
\newblock In \emph{Proceedings of the 2012 Conference of the North American
  Chapter of the Association for Computational Linguistics (NAACL 2012)}, pages
  568--572.

\bibitem[{Dahlmeier et~al.(2013)Dahlmeier, Ng, and Wu}]{Dahlmeier:13:BEA}
Daniel Dahlmeier, Hwee~Tou Ng, and Siew~Mei Wu. 2013.
\newblock {Building a Large Annotated Corpus of Learner English: The NUS Corpus
  of Learner English}.
\newblock In \emph{Proceedings of the 8th Workshop on Building Educational
  Applications Using NLP (BEA 2013)}, pages 22--31.

\bibitem[{Devlin et~al.(2019)Devlin, Chang, Lee, and
  Toutanova}]{devlin-etal-2019-bert}
Jacob Devlin, Ming-Wei Chang, Kenton Lee, and Kristina Toutanova. 2019.
\newblock {{BERT}: Pre-training of Deep Bidirectional Transformers for Language
  Understanding}.
\newblock In \emph{Proceedings of the 2019 Conference of the North {A}merican
  Chapter of the Association for Computational Linguistics (NAACL 2019)}, pages
  4171--4186.

\bibitem[{Felice and Pulman(2008)}]{Felice:08:COLING}
Rachele~De Felice and Stephen~G. Pulman. 2008.
\newblock {A Classifier-Based Approach to Preposition and Determiner Error
  Correction in {L}2 {E}nglish}.
\newblock In \emph{Proceedings of the 22nd International Conference on
  Computational Linguistics (COLING 2008)}, pages 169--176.

\bibitem[{Ge et~al.(2018)Ge, Wei, and Zhou}]{ge2018reaching}
Tao Ge, Furu Wei, and Ming Zhou. 2018.
\newblock {Reaching Human-level Performance in Automatic Grammatical Error
  Correction: An Empirical Study}.
\newblock \emph{arXiv preprint arXiv:1807.01270}.

\bibitem[{Geertzen et~al.(2013)Geertzen, Alexopoulou, and Korhonen}]{efcamdat}
Jeroen Geertzen, Dora Alexopoulou, and Anna Korhonen. 2013.
\newblock \emph{Automatic linguistic annotation of large scale L2 databases:
  The EF-Cambridge Open Language Database (EFCAMDAT)}.

\bibitem[{Granger(1998)}]{granger:1998:LEC}
Sylviane Granger. 1998.
\newblock {The computer learner corpus: A versatile new source of data for SLA
  research}.
\newblock In Sylviane Granger, editor, \emph{{Learner English on Computer}},
  pages 3--18.

\bibitem[{Grundkiewicz et~al.(2019)Grundkiewicz, Junczys-Dowmunt, and
  Heafield}]{grundkiewicz:2019:bea}
Roman Grundkiewicz, Marcin Junczys-Dowmunt, and Kenneth Heafield. 2019.
\newblock {Neural Grammatical Error Correction Systems with Unsupervised
  Pre-training on Synthetic Data}.
\newblock In \emph{Proceedings of the Fourteenth Workshop on Innovative Use of
  NLP for Building Educational Applications (BEA 2019)}, pages 252--263.

\bibitem[{Han et~al.(2006)Han, Chodorow, and Leacock}]{Han:06:Journal}
Na-Rae Han, Martin Chodorow, and Claudia Leacock. 2006.
\newblock {Detecting Errors in English Article Usage by Non-Native Speakers}.
\newblock \emph{Natural Language Engineering}, 12(2):115--129.

\bibitem[{He et~al.(2020)He, Gu, Shen, and Ranzato}]{He2020Revisiting}
Junxian He, Jiatao Gu, Jiajun Shen, and Marc'Aurelio Ranzato. 2020.
\newblock {Revisiting Self-Training for Neural Sequence Generation}.
\newblock In \emph{International Conference on Learning Representations (ICLR
  2020)}.

\bibitem[{Junczys-Dowmunt(2018)}]{junczys-dowmunt2018wmt:filtering}
Marcin Junczys-Dowmunt. 2018.
\newblock {Dual Conditional Cross-Entropy Filtering of Noisy Parallel Corpora}.
\newblock In \emph{Proceedings of the Third Conference on Machine Translation:
  Shared Task Papers (WMT 2018)}, pages 888--895.

\bibitem[{Junczys-Dowmunt et~al.(2018)Junczys-Dowmunt, Grundkiewicz, Guha, and
  Heafield}]{junczys:2018:NAACL}
Marcin Junczys-Dowmunt, Roman Grundkiewicz, Shubha Guha, and Kenneth Heafield.
  2018.
\newblock {Approaching Neural Grammatical Error Correction as a Low-Resource
  Machine Translation Task}.
\newblock In \emph{Proceedings of the 2018 Conference of the North American
  Chapter of the Association for Computational Linguistics (NAACL 2018)}, pages
  595--606.

\bibitem[{Kaneko et~al.(2020)Kaneko, Mita, Kiyono, Suzuki, and
  Inui}]{Kaneko2020EncoderDecoderMC}
Masahiro Kaneko, Masato Mita, Shun Kiyono, Jun Suzuki, and Kentaro Inui. 2020.
\newblock {Encoder-Decoder Models Can Benefit from Pre-trained Masked Language
  Models in Grammatical Error Correction}.
\newblock \emph{arXiv preprint arXiv:2005.00987}.

\bibitem[{Khayrallah and Koehn(2018)}]{khayrallah-koehn-2018-impact}
Huda Khayrallah and Philipp Koehn. 2018.
\newblock {On the Impact of Various Types of Noise on Neural Machine
  Translation}.
\newblock In \emph{Proceedings of the 2nd Workshop on Neural Machine
  Translation and Generation}, pages 74--83.

\bibitem[{Kingma and Ba(2015)}]{kingma:2015:ICLR}
Diederik Kingma and Jimmy Ba. 2015.
\newblock {Adam: A Method for Stochastic Optimization}.
\newblock In \emph{Proceedings of the 3rd International Conference on Learning
  Representations (ICLR 2015)}.

\bibitem[{Kiyono et~al.(2019)Kiyono, Suzuki, Mita, Mizumoto, and
  Inui}]{kiyono-etal-2019-empirical}
Shun Kiyono, Jun Suzuki, Masato Mita, Tomoya Mizumoto, and Kentaro Inui. 2019.
\newblock {An Empirical Study of Incorporating Pseudo Data into Grammatical
  Error Correction}.
\newblock In \emph{Proceedings of the 2019 Conference on Empirical Methods in
  Natural Language Processing and the 9th International Joint Conference on
  Natural Language Processing (EMNLP-IJCNLP 2019)}, pages 1236--1242.

\bibitem[{Koehn et~al.(2018)Koehn, Khayrallah, Heafield, and
  Forcada}]{koehn-etal-2018-findings}
Philipp Koehn, Huda Khayrallah, Kenneth Heafield, and Mikel~L. Forcada. 2018.
\newblock {Findings of the {WMT} 2018 Shared Task on Parallel Corpus
  Filtering}.
\newblock In \emph{Proceedings of the Third Conference on Machine Translation
  (WMT 2018): Shared Task Papers}, pages 726--739.

\bibitem[{Koehn and Knowles(2017)}]{koehn:2017:NMT}
Philipp Koehn and Rebecca Knowles. 2017.
\newblock {Six Challenges for Neural Machine Translation}.
\newblock In \emph{Proceedings of the First Workshop on Neural Machine
  Translation (WMT 2018)}, pages 28--39.

\bibitem[{Lichtarge et~al.(2020)Lichtarge, Alberti, and
  Kumar}]{lichtarge2020data}
Jared Lichtarge, Chris Alberti, and Shankar Kumar. 2020.
\newblock Data weighted training strategies for grammatical error correction.
\newblock \emph{arXiv preprint arXiv:2008.02976}.

\bibitem[{Lichtarge et~al.(2019)Lichtarge, Alberti, Kumar, Shazeer, Parmar, and
  Tong}]{lichtarge-etal-2019-corpora}
Jared Lichtarge, Chris Alberti, Shankar Kumar, Noam Shazeer, Niki Parmar, and
  Simon Tong. 2019.
\newblock {Corpora Generation for Grammatical Error Correction}.
\newblock In \emph{Proceedings of the 2019 Conference of the North American
  Chapter of the Association for Computational Linguistics (NAACL 2019)}, pages
  3291--3301.

\bibitem[{Liu et~al.(2019)Liu, Ott, Goyal, Du, Joshi, Chen, Levy, Lewis,
  Zettlemoyer, and Stoyanov}]{Liu2019RoBERTaAR}
Yinhan Liu, Myle Ott, Naman Goyal, Jingfei Du, Mandar Joshi, Danqi Chen, Omer
  Levy, Mike Lewis, Luke Zettlemoyer, and Veselin Stoyanov. 2019.
\newblock {RoBERTa: A Robustly Optimized BERT Pretraining Approach}.
\newblock \emph{arXiv preprint arXiv:1907.11692}.

\bibitem[{Lo et~al.(2018)Lo, Chen, Yang, and Chang}]{lo-etal-2018-cool}
Yu-Chun Lo, Jhih-Jie Chen, Chingyu Yang, and Jason Chang. 2018.
\newblock {Cool {E}nglish: a Grammatical Error Correction System Based on Large
  Learner Corpora}.
\newblock In \emph{Proceedings of the 27th International Conference on
  Computational Linguistics (COLING 2018): System Demonstrations}, pages
  82--85.

\bibitem[{Mita et~al.(2019)Mita, Mizumoto, Kaneko, Nagata, and
  Inui}]{mita-etal-2019-cross}
Masato Mita, Tomoya Mizumoto, Masahiro Kaneko, Ryo Nagata, and Kentaro Inui.
  2019.
\newblock {Cross-Corpora Evaluation and Analysis of Grammatical Error
  Correction Models {---} Is Single-Corpus Evaluation Enough?}
\newblock In \emph{Proceedings of the 2019 Conference of the North {A}merican
  Chapter of the Association for Computational Linguistics (NAACL 2019)}, pages
  1309--1314.

\bibitem[{Mizumoto et~al.(2011)Mizumoto, Komachi, Nagata, and
  Matsumoto}]{mizumoto:2011:IJCNLP}
Tomoya Mizumoto, Mamoru Komachi, Masaaki Nagata, and Yuji Matsumoto. 2011.
\newblock {Mining Revision Log of Language Learning SNS for Automated Japanese
  Error Correction of Second Language Learners}.
\newblock In \emph{Proceedings of the 5th International Joint Conference on
  Natural Language Processing (IJCNLP 2011)}, pages 147--155.

\bibitem[{Nagata et~al.(2006)Nagata, Kawai, Morihiro, and
  Isu}]{Nagata:06:COLING}
Ryo Nagata, Atsuo Kawai, Koichiro Morihiro, and Naoki Isu. 2006.
\newblock A feedback-augmented method for detecting errors in the writing of
  learners of {E}nglish.
\newblock In \emph{Proceedings of the 21st International Conference on
  Computational Linguistics and 44th Annual Meeting of the Association for
  Computational Linguistics (COLING-ACL 2006)}, pages 241--248.

\bibitem[{Napoles et~al.(2015)Napoles, Sakaguchi, Post, and
  Tetreault}]{napoles:2015:ACL}
Courtney Napoles, Keisuke Sakaguchi, Matt Post, and Joel Tetreault. 2015.
\newblock {Ground Truth for Grammatical Error Correction Metrics}.
\newblock In \emph{Proceedings of the 53rd Annual Meeting of the Association
  for Computational Linguistics and the 7th International Joint Conference on
  Natural Language Processing (ACL \& IJCNLP 2015)}, pages 588--593.

\bibitem[{Napoles et~al.(2016)Napoles, Sakaguchi, Post, and
  Tetreault}]{napoles:2016:gleu}
Courtney Napoles, Keisuke Sakaguchi, Matt Post, and Joel Tetreault. 2016.
\newblock {GLEU Without Tuning}.
\newblock \emph{arXiv preprint arXiv:1605.02592}.

\bibitem[{Napoles et~al.(2017)Napoles, Sakaguchi, and
  Tetreault}]{Napoles:17:EACL}
Courtney Napoles, Keisuke Sakaguchi, and Joel Tetreault. 2017.
\newblock {JFLEG: A Fluency Corpus and Benchmark for Grammatical Error
  Correction}.
\newblock In \emph{Proceedings of the 15th Conference of the European Chapter
  of the Association for Computational Linguistics (EACL 2017)}, pages
  229--234.

\bibitem[{Nie et~al.(2019)Nie, Yao, Wang, Pan, and Lin}]{nie-etal-2019-simple}
Feng Nie, Jin-Ge Yao, Jinpeng Wang, Rong Pan, and Chin-Yew Lin. 2019.
\newblock A simple recipe towards reducing hallucination in neural surface
  realisation.
\newblock In \emph{Proceedings of the 57th Annual Meeting of the Association
  for Computational Linguistics (ACL 2019)}, pages 2673--2679.

\bibitem[{Omelianchuk et~al.(2020)Omelianchuk, Atrasevych, Chernodub, and
  Skurzhanskyi}]{omelianchuk2020gector}
Kostiantyn Omelianchuk, Vitaliy Atrasevych, Artem Chernodub, and Oleksandr
  Skurzhanskyi. 2020.
\newblock {GECToR -- Grammatical Error Correction: Tag, Not Rewrite}.
\newblock \emph{arXiv preprint arXiv:2005.12592}.

\bibitem[{Ott et~al.(2019)Ott, Edunov, Baevski, Fan, Gross, Ng, Grangier, and
  Auli}]{ott2019:arxiv:fairseq}
Myle Ott, Sergey Edunov, Alexei Baevski, Angela Fan, Sam Gross, Nathan Ng,
  David Grangier, and Michael Auli. 2019.
\newblock {fairseq: A Fast, Extensible Toolkit for Sequence Modeling}.
\newblock In \emph{Proceedings of the 2019 Conference of the North American
  Chapter of the Association for Computational Linguistics (NAACL 2019)}, pages
  48--53.

\bibitem[{Radford et~al.(2019)Radford, Wu, Child, Luan, Amodei, and
  Sutskever}]{radford2019language}
Alec Radford, Jeff Wu, Rewon Child, David Luan, Dario Amodei, and Ilya
  Sutskever. 2019.
\newblock Language models are unsupervised multitask learners.

\bibitem[{Rossenbach et~al.(2018)Rossenbach, Rosendahl, Kim, Gra{\c{c}}a,
  Gokrani, and Ney}]{rossenbach-etal-2018-rwth}
Nick Rossenbach, Jan Rosendahl, Yunsu Kim, Miguel Gra{\c{c}}a, Aman Gokrani,
  and Hermann Ney. 2018.
\newblock {The {RWTH} Aachen University Filtering System for the {WMT} 2018
  Parallel Corpus Filtering Task}.
\newblock In \emph{Proceedings of the Third Conference on Machine Translation
  (WMT 2018): Shared Task Papers}, pages 946--954.

\bibitem[{Sennrich et~al.(2017)Sennrich, Birch, Currey, Germann, Haddow,
  Heafield, Miceli~Barone, and Williams}]{sennrich:2017:wmt}
Rico Sennrich, Alexandra Birch, Anna Currey, Ulrich Germann, Barry Haddow,
  Kenneth Heafield, Antonio~Valerio Miceli~Barone, and Philip Williams. 2017.
\newblock The university of {E}dinburgh{'}s neural {MT} systems for {WMT}17.
\newblock In \emph{Proceedings of the Second Conference on Machine Translation
  (WMT 2017)}, pages 389--399.

\bibitem[{Sennrich et~al.(2016{\natexlab{a}})Sennrich, Haddow, and
  Birch}]{sennrich:2016:wmt}
Rico Sennrich, Barry Haddow, and Alexandra Birch. 2016{\natexlab{a}}.
\newblock {E}dinburgh neural machine translation systems for {WMT} 16.
\newblock In \emph{Proceedings of the First Conference on Machine Translation:
  Volume 2, Shared Task Papers (WMT 2016)}, pages 371--376.

\bibitem[{Sennrich et~al.(2016{\natexlab{b}})Sennrich, Haddow, and
  Birch}]{sennrich:2016:ACL}
Rico Sennrich, Barry Haddow, and Alexandra Birch. 2016{\natexlab{b}}.
\newblock {Neural Machine Translation of Rare Words with Subword Units}.
\newblock In \emph{Proceedings of the 54th Annual Meeting of the Association
  for Computational Linguistics (ACL 2016)}, pages 1715--1725.

\bibitem[{Sennrich and Zhang(2019)}]{sennrich:2019:ACL}
Rico Sennrich and Biao Zhang. 2019.
\newblock {Revisiting Low-Resource Neural Machine Translation: A Case Study}.
\newblock In \emph{Proceedings of the 57th Annual Meeting of the Association
  for Computational Linguistics (ACL 2019)}, pages 211--221.

\bibitem[{Sutskever et~al.(2014)Sutskever, Vinyals, and
  Le}]{sutskever:2014:NIPS}
Ilya Sutskever, Oriol Vinyals, and Quoc~V. Le. 2014.
\newblock {Sequence to Sequence Learning with Neural Networks}.
\newblock In \emph{Advances in Neural Information Processing Systems 28 (NIPS
  2014)}, pages 3104--3112.

\bibitem[{Szegedy et~al.(2016)Szegedy, Vanhoucke, Ioffe, Shlens, and
  Wojna}]{szegedy:2016:rethinking}
Christian Szegedy, Vincent Vanhoucke, Sergey Ioffe, Jon Shlens, and Zbigniew
  Wojna. 2016.
\newblock Rethinking the inception architecture for computer vision.
\newblock In \emph{2016 IEEE Conference on Computer Vision and Pattern
  Recognition (CVPR 2016)}, pages 2818--2826.

\bibitem[{Tajiri et~al.(2012)Tajiri, Komachi, and Matsumoto}]{Tajiri:12:ACL}
Toshikazu Tajiri, Mamoru Komachi, and Yuji Matsumoto. 2012.
\newblock {Tense and Aspect Error Correction for ESL Learners Using Global
  Context}.
\newblock In \emph{Proceedings of the 50th Annual Meeting of the Association
  for Computational Linguistics (ACL 2012)}, pages 198--202.

\bibitem[{Vaswani et~al.(2017)Vaswani, Shazeer, Parmar, Uszkoreit, Jones,
  Gomez, Kaiser, and Polosukhin}]{vaswani:2017:NIPS}
Ashish Vaswani, Noam Shazeer, Niki Parmar, Jakob Uszkoreit, Llion Jones,
  Aidan~N Gomez, {\L}ukasz Kaiser, and Illia Polosukhin. 2017.
\newblock {Attention Is All You Need}.
\newblock In \emph{Advances in Neural Information Processing Systems 31 (NIPS
  2017)}, pages 5998--6008.

\bibitem[{Wang(2019)}]{wang-2019-revisiting}
Hongmin Wang. 2019.
\newblock Revisiting challenges in data-to-text generation with fact grounding.
\newblock In \emph{Proceedings of the 12th International Conference on Natural
  Language Generation (INLG 2019)}, pages 311--322.

\bibitem[{Xie et~al.(2020)Xie, Luong, Hovy, and Le}]{Xie_2020_CVPR}
Qizhe Xie, Minh-Thang Luong, Eduard Hovy, and Quoc~V. Le. 2020.
\newblock Self-training with noisy student improves imagenet classification.
\newblock In \emph{Proceedings of the IEEE/CVF Conference on Computer Vision
  and Pattern Recognition (CVPR 2020)}.

\bibitem[{Yang et~al.(2019)Yang, Dai, Yang, Carbonell, Salakhutdinov, and
  Le}]{xlnet-nips2019}
Zhilin Yang, Zihang Dai, Yiming Yang, Jaime Carbonell, Russ~R Salakhutdinov,
  and Quoc~V Le. 2019.
\newblock {XLNet: Generalized Autoregressive Pretraining for Language
  Understanding}.
\newblock In H.~Wallach, H.~Larochelle, A.~Beygelzimer, F.~d¥textquotesingle
  Alch¥'{e}-Buc, E.~Fox, and R.~Garnett, editors, \emph{Advances in Neural
  Information Processing Systems 32 (NIPS 2019)}, pages 5753--5763.

\bibitem[{Yannakoudakis et~al.(2018)Yannakoudakis, Andersen, Geranpayeh,
  Briscoe, and Nicholls}]{Yannakoudakis:18:Journal}
Helen Yannakoudakis, {\O}istein~E. Andersen, Ardeshir Geranpayeh, Ted Briscoe,
  and Diane Nicholls. 2018.
\newblock {Developing an Automated Writing Placement system for ESL Learners}.
\newblock \emph{Applied Measurement in Education}, 31(3):251--267.

\bibitem[{Yannakoudakis et~al.(2011)Yannakoudakis, Briscoe, and
  Medlock}]{Yannakoudakis:11:ACL}
Helen Yannakoudakis, Ted Briscoe, and Ben Medlock. 2011.
\newblock {A New Dataset and Method for Automatically Grading ESOL Texts}.
\newblock In \emph{Proceedings of the 49th Annual Meeting of the Association
  for Computational Linguistics (ACL 2011)}, pages 180--189.

\bibitem[{Zhao et~al.(2019)Zhao, Wang, Shen, Jia, and Liu}]{zhao2019improving}
Wei Zhao, Liang Wang, Kewei Shen, Ruoyu Jia, and Jingming Liu. 2019.
\newblock {Improving Grammatical Error Correction via Pre-Training a
  Copy-Augmented Architecture with Unlabeled Data}.
\newblock In \emph{Proceedings of the 2019 Conference of the North American
  Chapter of the Association for Computational Linguistics (NAACL 2019)}, pages
  156--165.

\end{thebibliography}
\bibliographystyle{acl_natbib}

\newpage
\appendix

\newpage

\section{The BEA-2019 official dataset}
\label{appendix:bea-2019-dataset}
The BEA-2019 Shared Task provided participants with the following datasets as official datasets: Lang-8~\citep{mizumoto:2011:IJCNLP,Tajiri:12:ACL}, the National University of Singapore Corpus of Learner English (NUCLE)~\citep{Dahlmeier:13:BEA}, the First Certificate in English corpus~\citep{Yannakoudakis:11:ACL}, and W\&I+LOCNESS~\citep{Yannakoudakis:18:Journal,granger:1998:LEC}. 
The official dataset is publicly available at \url{https://www.cl.cam.ac.uk/research/nl/bea2019st/}.

\section{Hyper-parameter settings}
\label{appendix:hyper-parameter-settings}

\begin{table}[h]
\centering
\small
\begin{tabular}{@{}lp{40mm}@{}}
\toprule
 Configurations & Values \\ \midrule
 Model Architecture & Transformer~\citep{vaswani:2017:NIPS} \\
 Optimizer& Adam~\citep{kingma:2015:ICLR}\\
 Learning Rate Schedule & Same as described in Section 5.3 of \citet{vaswani:2017:NIPS} \\
 Number of Epochs & 30 \\
 Dropout& 0.3\\
 Stopping Criterion & Train model for 30 epochs. During the training, save model parameter for every 500 updates. \\ 
 Gradient Clipping& 1.0\\
 Loss Function& Label smoothed cross entropy~\citep{szegedy:2016:rethinking} \\
 Beam Search& Beam size 5 with length normalization\\
 \bottomrule
\end{tabular}
\caption{Detailed hyper-parameters used for the base GEC model.}
\end{table}

\section{Preliminary experiment of the cross-entropy filtering}
\label{appendix:cross-entorpy_exp}
We investigated the effectiveness of changing the threshold of CE filtering by evaluating the model performance on BEA-valid. 
In this study, we prepared a forward and reverse pre-train model using BEA-train and CoNLL-2013 for as a training and validation set, respectively.

\begin{figure}[h]
 \centering
\includegraphics[width=1.0\linewidth]{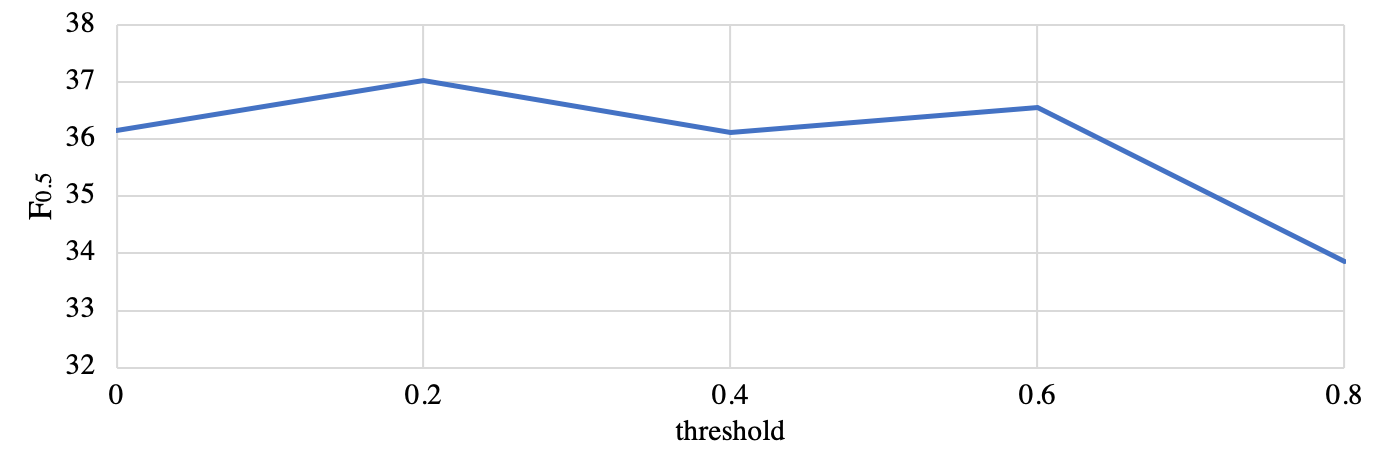}
 \caption{Performance of base GEC model on BEA-valid as threshold of CE filtering is varied.}
 \label{fig:ce_filter}
\end{figure}

\section{Examples of a confusion set before and after denoising}
\label{appendix:confusion_set}
Table~\ref{tab:conf} provides examples of a confusion set before and after applying the denoising method to EFCamDat.
We confirmed that we succeeded in reducing the noisy confusion set, including (*discuss about, *discuss about) or (*enter in, *enter in) in the target sentences using the proposed denoising.

\begin{table}[th]
\centering
\scriptsize
\begin{tabular}{lrr}
\toprule
Confusion set ($\bm{X},\bm{Y}$)& $\bm{Y}$ (\%) & $\bm{\hat{Y}}$ (\%)\\ \midrule
(*discuss about, *discuss about) & 66.7& 49.5\\
(*discuss about, discuss) & 33.0 & 50.2 \\ 
(*discuss about, *discuss in)& 0.3 & 0.3\\ \midrule 
(*enter in, *enter in)& 61.6& 31.7 \\
(*enter in,enter)& 38.4 & 68.3 \\ \bottomrule
\end{tabular}

 \caption{Examples of confusion set before and after denoising in EFCamDat.}
\label{tab:conf}
\end{table}

\section{Results of comparison with existing models}
\label{appendix:top_model}

\begin{table*}[h]
\centering
\scriptsize{}
\tabcolsep 1.5mm
\begin{tabular}{lccccccccccc}
\toprule
& \multicolumn{4}{c}{\begin{tabular}{c} CoNLL-2014 \end{tabular}} & \multicolumn{4}{c}{\begin{tabular}{c} JFLEG \end{tabular}}& \multicolumn{3}{c}{\begin{tabular}{c} BEA-test\end{tabular}} \\
\cmidrule(r){2-5}\cmidrule(r){6-9}\cmidrule(r){10-12}
\multicolumn{1}{c}{Model} & Prec. & Rec. & \fscore{}& GLEU& Prec. & Rec. & \fscore{}& GLEU & Prec. & Rec. & \fscore{} \\
\midrule
\scriptsize{\textbf{Single model:}}\\
\citet{junczys:2018:NAACL} & -& - & 53.0& -& - & - & - & 57.9 & -& - & -\\
\citet{lichtarge-etal-2019-corpora} & 65.5& 37.1 & 56.8 & -& - & - & - & 61.6 & -& - & -\\
\citet{awasthi-etal-2019-parallel} & 66.1& 43.0 & 59.7 & -& - & - & - & 60.3 & -& - & -\\
\citet{kiyono-etal-2019-empirical}& 67.9 &44.1& 61.3 & 68.6& \bf 76.6 & 55.8 & 71.3 & 59.7 & 65.5& 59.4 & 64.2 \\
\citet{Kaneko2020EncoderDecoderMC}& 69.2& 45.6 & 62.6& - & - & - & -& 61.3 &67.1& 60.1 & 65.6\\
\citet{omelianchuk2020gector}& \bf 77.5& 40.1 & \bf 65.3& - & - & - & -& - & \bf 79.2& 53.9 & \bf72.4\\
 SR $_{\textbf{BEA}+\textbf{EF}+\textbf{L8}}$+\textsc{PRET}& 63.8& \textbf{52.4} &61.1& \bf69.6& 74.9 & \textbf{62.5} & \bf72.0 & \textbf{63.4} & 59.9& \textbf{66.9} & 61.2 \\ 
\midrule
\scriptsize{\textbf{Ensemble model:}} \\
 \citet{junczys:2018:NAACL} & 61.9& 40.2 & 55.8 & -& - & - & -& 59.9& -& - & -\\
\citet{lichtarge-etal-2019-corpora} & 66.7& 43.9 & 60.4& -& - & - & - & 63.3& -& - & -\\
 \citet{grundkiewicz:2019:bea} & - &-& 64.2 & - & - & - & - & 61.2 & 72.3 & 60.1 & 69.5 \\
\citet{kiyono-etal-2019-empirical}& 72.4 &46.1& 65.0 & 68.8& \bf 79.5 & 54.6 & 72.9 & 61.4 & 74.7& 56.7 & 70.2 \\
\citet{Kaneko2020EncoderDecoderMC} &72.6 &46.4& 65.2& -& - & - & -& 62.0 & 72.3 & 61.4 & 69.8 \\
\citet{omelianchuk2020gector} &\bf 78.2 &41.5& \textbf{66.5}& -& - & - & -& - & \bf78.9 & 58.2 & \bf73.6 \\
 SR $_{\textbf{BEA}+\textbf{EF}+\textbf{L8}}$+\textsc{PRET} + \textsc{R2L} & 65.5 &\textbf{53.2}& 62.6 & \bf70.1 & 76.5 & \textbf{63.3} & \bf73.4 & \textbf{63.9}& 62.9 & \textbf{67.7} & 63.8\\

\bottomrule
\end{tabular}
\vskip -2mm
\caption{Comparison with existing models: a \textbf{bold} value denotes the best result within the column. SR and BEA indicate SR $_{\textbf{BEA}+\textbf{EF}+\textbf{L8}}$ and BEA-test, respectively. \citet{Kaneko2020EncoderDecoderMC} and \citet{omelianchuk2020gector} have appeared on arXiv less than 3 months before our submission and are considered contemporaneous to our submission. } 
\label{tab:sota}
\end{table*}

\section{Ablation study of SED}
\label{appendix:ablation_sed}

\begin{table*}[h]
\centering
\scriptsize{}
\tabcolsep 1.5mm
\begin{tabular}{lccccccccccc}
\toprule
& \multicolumn{4}{c}{\begin{tabular}{c} CoNLL-2014 \end{tabular}} & \multicolumn{4}{c}{\begin{tabular}{c} JFLEG \end{tabular}}& \multicolumn{3}{c}{\begin{tabular}{c} BEA-test\end{tabular}} \\
\cmidrule(r){2-5}\cmidrule(r){6-9}\cmidrule(r){10-12}
\multicolumn{1}{c}{Model} & Prec. & Rec. & \fscore{}& GLEU& Prec. & Rec. & \fscore{}& GLEU & Prec. & Rec. & \fscore{} \\
\midrule
\scriptsize{\textbf{Single model:}}\\
 SR $_{\textbf{BEA}+\textbf{EF}+\textbf{L8}}$+\textsc{PRET}& 63.8& \textbf{52.4} &61.1& 69.6& 74.9 & \textbf{62.5} & 72.0 & \textbf{63.4} & 59.9& \textbf{66.9} & 61.2 \\ 
SR $_{\textbf{BEA}+\textbf{EF}+\textbf{L8}}$ +\textsc{PRET}+\textsc{SED} & 65.2& \bf 49.9 &61.4 & \textbf{69.3} & 76.3 &\bf 60.6 & \textbf{72.5} &\bf 63.3 & 66.7&\bf 61.3 & 65.5 \\ 
\midrule
\scriptsize{\textbf{Ensemble model:}} \\
 SR $_{\textbf{BEA}+\textbf{EF}+\textbf{L8}}$+\textsc{PRET} + \textsc{R2L} & 65.5 &\textbf{53.2}& 62.6 & 70.1 & 76.5 & \textbf{63.3} & 73.4 & \textbf{63.9}& 62.9 & \textbf{67.7} & 63.8\\ 
 SR $_{\textbf{BEA}+\textbf{EF}+\textbf{L8}}$+\textsc{PRET}+\textsc{R2L}+\textsc{SED}& 67.1 &\bf 50.8& 63.1& \textbf{69.8}& 77.8 & \bf 61.5 & \textbf{73.9}&\bf 63.7& 69.4 &\bf 62.1 & 67.8\\ 

\bottomrule
\end{tabular}
\vskip -2mm
\caption{Ablation study of SED}
\label{tab:ablation_sed}
\end{table*}

\end{document}